\documentclass[letterpaper, 10 pt, conference]{ieeeconf}  

\IEEEoverridecommandlockouts                              

\usepackage{graphicx}
\usepackage{siunitx}
\usepackage{mathptmx} 
\usepackage{times} 
\usepackage{amsmath} 
\usepackage{amssymb}  
\usepackage[T1]{fontenc}
\usepackage{algorithm}
\usepackage{algorithmic}
\usepackage{url}
\usepackage{booktabs}

\title{\LARGE \bf
Spline-Based Minimum-Curvature Trajectory Optimization for Autonomous Racing
}

\author{Haoru Xue$^{1}$, Tianwei Yue$^{2}$ and John M. Dolan$^{1}$
\thanks{${^{1}}$ Haoru Xue and John M. Dolan are with the Robotics Institute, Carnegie Mellon University {\tt\small \{haorux, jdolan\}@andrew.cmu.edu}}%
\thanks{${^{2}}$ Tianwei Yue is with the Language Technologies Institute, Carnegie Mellon University {\tt\small tyue@andrew.cmu.edu}}%
\thanks{${^{3}}$ Codes: https://github.com/HaoruXue/spline-trajectory-optimization}%
}

\begin{document}

\maketitle
\thispagestyle{empty}
\pagestyle{empty}

\begin{abstract}

We propose a novel B-spline trajectory optimization method for autonomous racing. We consider the unavailability of sophisticated race car and race track dynamics in early-stage autonomous motorsports development and derive methods that work with limited dynamics data and additional conservative constraints. We formulate a minimum-curvature optimization problem with only the spline control points as optimization variables. We then compare the current state-of-the-art method with our optimization result, which achieves a similar level of optimality with a 90\% reduction on the decision variable dimension, and in addition offers mathematical smoothness guarantee and flexible manipulation options. We concurrently reduce the problem computation time from seconds to milliseconds for a long race track, enabling future online adaptation of the previously offline technique.

\end{abstract}

\section{INTRODUCTION}

\subsection{Offline Trajectory Optimization in Autonomous Racing}

Offline trajectory optimization (OTO) is widely used in modern autonomous racing. By leveraging sophisticated prior knowledge of the race track (geometries, friction conditions, etc.) and knowledge of the race car (tire model, power train, etc.), an optimization program can be run in a reasonable time frame to achieve best lap time. The result can then significantly reduce the online computation load in sample-based planning \cite{werling_optimal_2010} and model predictive control \cite{novi_real-time_2020}.

Recent advancements in high-speed autonomous racing present new challenges and opportunities for evaluating OTO algorithms. Lack of prior race car and race track data is a significant challenge for university-level research and racing development. For example, although autonomous race cars in the Indy Autonomous Challenge (IAC) have reached a top speed of over 320 km/h, the teams still have limited access to critical data such as tire model parameters and load transfer characteristics, especially in the earlier stages of development, when estimation of these parameters is not viable with the limited data gathered. Therefore, a simple trajectory optimization algorithm should account for this data-scarce use case and support early development efforts.

OTO is often used to generate a reference safety set in the development phase of an autonomous race car, which is often desired when the handling limit of the vehicle is yet to be determined. Instead of directly applying an experimental tire model as the optimization limit, it is often desired to work with more conservative handling constraints. The traction circle (ellipse, or diamond) is an intuitive and effective alternative to the raw tire parameters. By controlling the maximum acceptable lateral and longitudinal acceleration in each direction as hyper-parameters, an OTO approach can generate different trajectories subjected to additional dynamics constraints. In addition, the track geometry constraints can be altered to impose extra track limits, and generate trajectories that pass through the non-optimal part of the race track, which provides a planning and control reference when the vehicle is forced into these regions.

\subsection{Related Work}

Prior to autonomous racing, the generation of an optimal velocity profile given a fixed trajectory was first studied in the motorsports domain. Quasi-steady state (QSS) approaches have been developed since the 1980s
\cite{brayshaw_quasi_2005}\cite{brayshaw_use_2005}\cite{cambiaghi_tool_nodate}\cite{siegler_lap_2000}. The full trajectory is broken down into small segments, through which the race car is assumed to have steady-state behavior. The algorithm starts with segments corresponding to peak curvature on the trajectory, which are considered the "bottlenecks". The algorithm then proceeds to generate a full velocity profile entering and exiting these bottlenecks, considering only the neighboring vehicle states subject to the dynamic limits, until the velocity profiles of these bottlenecks meet each other \cite{dal_bianco_comparison_2019}. This method is known for its robustness and fast run time, but it fails to capture transient effects such as load transfer characteristics and damper dynamics \cite{dal_bianco_comparison_2019}. We adopt a similar method in our work, whose algorithm will be formally proposed in a later section.

The optimization of the trajectory geometry in autonomous racing has been studied in Braghin et al. and Kapania et al. with the "minimum curvature" heuristic, which states that an optimal racing trajectory minimizes the sum of curvatures around the track to minimize lap time \cite{braghin_race_2008} \cite{kapania_sequential_2016}. Heilmeier et al. extend the idea to a quadratic programming (QP) formulation with improvements to the curvature calculation. They also apply spline interpolation to the noisy raw data and the final output to obtain a smooth trajectory \cite{heilmeier_minimum_2020}. Their work was extensively used by the TUM team in recent high-speed autonomous racing events such as Roborace and IAC. However, to guarantee that the continuity of the trajectory during and after optimization is at least $C^2$ (continuous position, velocity, and acceleration), it is insufficient to optimize with respect to discrete samples along the trajectory (3.0 m interval in \cite{heilmeier_minimum_2020}), although spline interpolation could be applied in postprocessing.

\subsection{Goal and Scope of This Work}

In our work, we propose a new optimization formulation for minimum-curvature OTO based on B-splines that ensures the continuity of the trajectory throughout the optimization iterations. We also consider the unavailability of sophisticated race car and race track dynamics in early-stage autonomous racing development, and derive methods that work with limited dynamics data and additional conservative constraints.

We introduce the math related to the B-spline and our cost function in Section \ref{background}. We then formulate the optimization problem in Section \ref{method}, and discuss the QSS algorithm to calculate the velocity profile, which is used to evaluate the generated trajectory. Finally, in Section \ref{results}, we compare the optimization results with previous works.

\section{BACKGROUND}\label{background}

A single B-spline of order $n$ is a parametrized curve, denoted as $B_{i,n}(x)$. It can be uniquely constructed from a series of nondescending knot points $t_0, t_1, \dots, t_N$, subjected to
\begin{align}
    \sum_{i=1}^{N-n}B_{i,n}(x) = 1
\end{align}

A B-spline has non-zero values only in the range of knot vectors $t_i < x \le t_{i+n}$. Higher-order B-splines can be recursively defined.
\begin{align}\label{eq:bspline}
    B_{i,n+1}(x)&=w_{i,n}(x)B_{i,n}(x)+(1-w_{i+1,n}(x))B_{i+1,n}(x)  \nonumber\\
    \text{where }
    w_{i,k}(x)&=\begin{cases}
      \frac{x-t_i}{t_{i+k}-t_i}, & t_{i+k}\ne t_i \\
      0, & \text{otherwise}
    \end{cases}
\end{align}

\begin{figure}[thpb]
      \centering
      \smallskip
      \smallskip
      \framebox{\parbox{3in}{\begin{center}
          \includegraphics[scale=0.2]{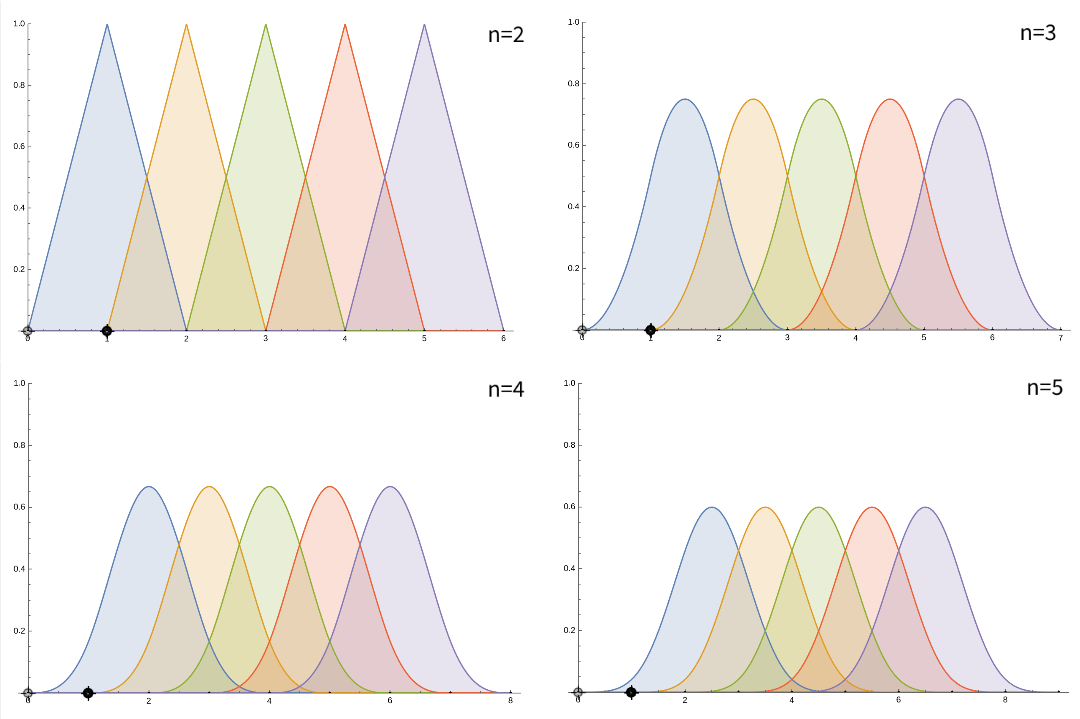}
      \end{center}}}
      \caption{Visualization of basis functions $B_{i,n}(x)$ of order 2 to 5 \cite{jan_dijkema_b-spline_2011}}
      \label{fig1}
\end{figure}

The resulting basis functions are $C^{n-2}$ continuous and overlap throughout the knot sequence, which is visualized in fig. \ref{fig1}.


These basis functions allow us to define a spline on $t_0, t_N$ that is a linear combination of the basis functions:
\begin{align}
    T_{n}(x)=\sum_{i=1}^{N-n}\alpha_iB_{i,n}(x)
\end{align}

\begin{figure}[thpb]
      \centering
      \includegraphics[scale=0.55,trim={0.5cm 0 0.5cm 1.26cm},clip]{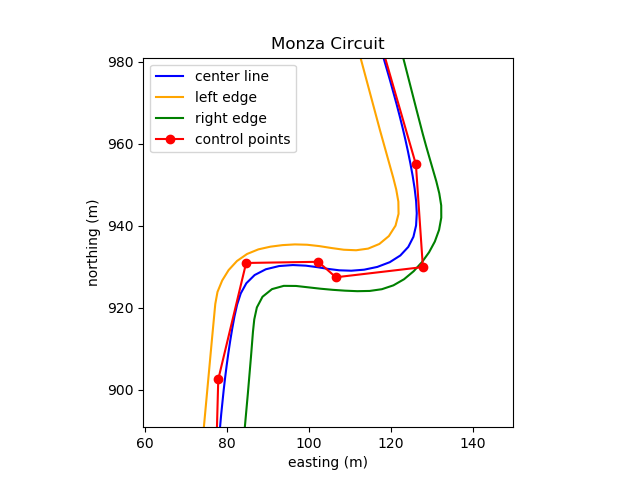}
      \caption{Turn 1 and 2 of Monza Circuit after interpolation}
      \label{fig2}
   \end{figure}

The weights $\alpha_0 \dots \alpha_{N-n}$ are also known as control points, which can be visualized in fig. \ref{fig2}. We use $S=N-n$ to denote the number of control points. The shape of the curve can be manipulated by moving the control points while keeping the basis functions constant. The movement of the control points is the main subject of interest in this work, and we aim to derive an optimization formulation that optimizes their placement to form a curvature-optimal trajectory for autonomous racing.

To extend the 1D B-spline to handle a trajectory in the 2D plane, we take two sets of control points to parameterize the $x$ and $y$ coordinates separately on the same basis functions. That is, given a trajectory $T(t): \mathbb{R} \rightarrow \mathbb{R}^2$ and a sequence of control points $\mathbf{z}=[\alpha_1,\dots,\alpha_{S},\beta_1,\dots,\beta_{S}]^T$
\begin{align}
    T(t, \mathbf{z})=(\sum_{i=1}^{S}\alpha_iB_{i,n}(t), \sum_{i=1}^{S}\beta_iB_{i,n}(t))
\end{align}
where $\alpha_i, \beta_i$ respectively denote the $x$ and $y$ coordinates of the control point. Conventionally, we use $t\in[0.0, 1.0]$ to parameterize the trajectory and denote progress along the track. We can also denote the two resulting 1D B-splines as $T_x(t. \mathbf{z}), T_y(t, \mathbf{z})$.

Since we will be optimizing with respect to the control points, it is useful to take the derivative of a spline with respect to the control points, which is simply the corresponding basis function.
\begin{align}
    \frac{\partial T(t,\mathbf{z})}{\partial \alpha_i} = \frac{\partial T(t,\mathbf{z})}{\partial \beta_i} = B_{i,n}(t)
\end{align}

The curvature of a B-spline trajectory, as of any parametric curve equation, can be calculated as \cite{heilmeier_minimum_2020}
\begin{align}
    \label{eq:curvature}
    k(t) = \frac{T_x'(t)T_y''(t) - T_y'(t)T_x''(t)}{(T_x'(t)^2+T_y'(t)^2)^{\frac{3}{2}}}
\end{align}

B-Splines have certain advantageous properties for trajectory optimization problems with respect to the control point movements. First, each control point can manipulate a piece of segment holistically. Intuitively, given the same knots \(t_0\dots t_n\), a control point has a longer influence range along the curve as the degree of the B-spline increases. Visually in Fig. \ref{fig1}, the basis function corresponding to that control point spans more intervals of knots. Second, since the basis function is evaluated to zero outside of its intervals according to \eqref{eq:bspline}, we can perform partial optimizations to a specific section of the trajectory without affecting the others. Lastly, the resulting trajectory from the control point movements is guaranteed to have the same order of continuity as the original trajectory since it is still a B-spline of the same order.

These features of B-splines make them suitable for autonomous racing applications. For a vehicle to have continuous velocity and acceleration profiles, the trajectory should be at least $C^2$ continuous, subject to additional curvature constraints since vehicle kinematics is not omnidirectional. To optimize the vehicle's trajectory through a specific segment of the race track, we can control the scope of the optimization by controlling the number of control points to be included in the optimization problem, and obtain results that are perhaps more locally optimized for a particular turn, or more globally optimized through a combination of turns.

\section{METHOD} \label{method}

\subsection{Generating and Evaluating a Spline Trajectory}

As an example, in this work we will use the Monza Circuit, which is a 5.8 km (3.6 mile) long race track used in the Indy Autonomous Challenge. It has a combination of long straights, high-speed turns, and chicanes. We will also show experiments on an other race track in a later section.

We obtain the race track geometries from satellite images and geographical surveys. We represent the track using a reference center line, with left and right offsets to denote the distances to the track boundaries at every waypoint. We then draw a cubic B-spline interpolation of the reference center line using the least square periodic spline interpolation algorithm discussed in \cite{dierckx_algorithms_1982}, which forms a closed-loop spline on $t\in [0.0, 1.0]$.

We then discretize the trajectory by taking waypoints at a constant 3-meter interval. This is done by performing a numerical integration to calculate the length of the trajectory:
\begin{align}
    L(t_{\min}, t_{\max}) = \int_{t_{\min}}^{t_{\max}}{T_x'(t)^2 + T_y'(t)^2dt}
\end{align}
and solving a subsequent root-finding problem to find $t_i$ such that the trajectory advances by 3 \si{\meter}.
\begin{align}
    t_i = \operatorname*{argmin}_{t_i} \quad {L(t_{i-1}, t_i) - 3}
\end{align}

We note that the optimization is done on the continuous spline, and these discretization points are simply used for sampling the curvature throughout the trajectory. The discrete trajectory is also useful in the evaluation process to be described below.

\subsection{Configuring the Vehicle Parameters}

\begin{figure}[thpb]
      \centering
      \includegraphics[scale=0.6,trim={0 0 0 1.35cm},clip]{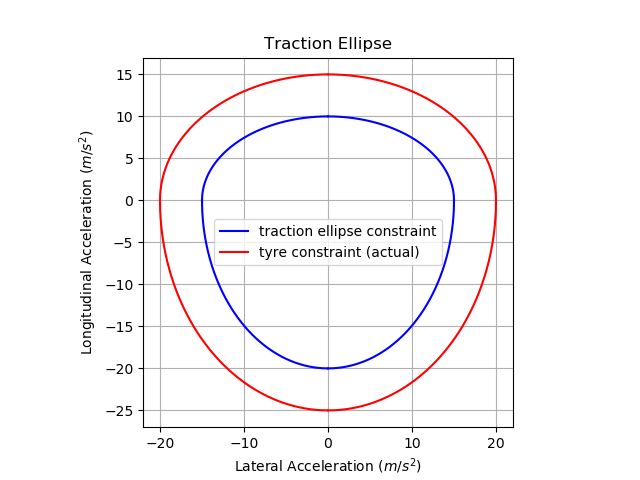}
      \caption{Geometry of traction ellipse used as constraint (blue), and actual tire constraints (red, simulated)}
      \label{fig3}
   \end{figure}

As the main constraint on vehicle dynamics, we use a traction ellipse, which considers the longitudinal and lateral accelerations of the vehicle when accelerating, braking and cornering, or a combination of them. To characterize the shape of the ellipse, four parameters are used, which correspond to the maximum longitudinal acceleration, longitudinal deceleration, and left and right lateral accelerations. For the purpose of autonomous racing development, these parameters are easily adjustable to impose additional safety constraints within tire limits. For race cars with asymmetrical setups, such as those racing on an oval racing circuit, the left and right lateral accelerations can be adjusted separately. In our example, shown in Fig. \ref{fig3}, we impose a maximum acceleration of 10 \si{\meter\per\second\squared}, deceleration of -20 \si{\meter\per\second\squared}, and symmetric maximum cornering load of $\pm 15$ \si{\meter\per\second\squared}.

\subsection{Simulating a Spline Trajectory}

We perform a simulation for the best lap time on the spline trajectory to obtain the velocity profile, which will be used in the evaluation. The QSS algorithm starts by examining the curvature profile of the discretized trajectory and using the minimum-curvature points as constraints for this simulation. The relationship between the vehicle's tangential velocity $v$ and curvature $k$ satisfies
\begin{align}
    v=\sqrt{a_{lat}/k}
\end{align}
where $a_{lat}$ is the lateral acceleration. Therefore, to maximize vehicle velocity through the bottleneck, the vehicle should have zero longitudinal acceleration to maximize lateral acceleration according to the traction ellipse.

\begin{figure}[thpb]
      \centering
      \smallskip
      \smallskip
      \includegraphics[scale=0.1333,trim={0.6cm 3.8cm 0.4cm 3.1cm},clip]{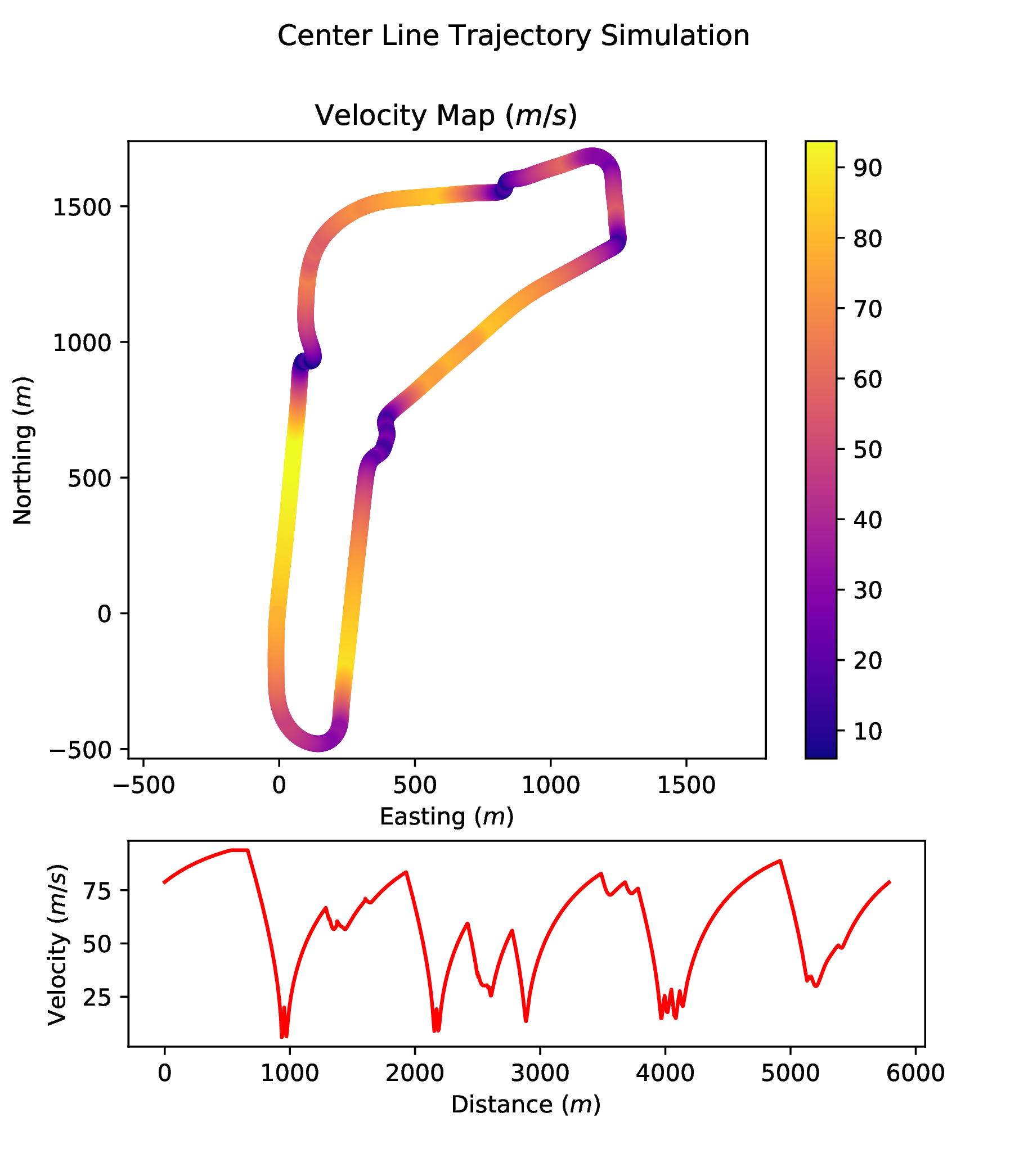}
      \caption{Example simulation on the center line trajectory. The heatmap shows the velocity levels (m/s) at individual track sections.}
      \label{fig4}
   \end{figure}

After obtaining this initial condition, the algorithm proceeds to generate an entry and exit velocity profile around the bottleneck points, until the two profiles of two bottlenecks converge. The trajectory is then further adjusted to ensure a smooth acceleration and velocity transition at the meeting points. Fig. \ref{fig4} shows a baseline simulation done on the reference center line trajectory, with a heatmap indicating the velocity levels that the vehicle can achieve in various sections of the track.


\subsection{Spline-Based Minimum Curvature Problem}

Building on \cite{braghin_race_2008} and \cite{heilmeier_minimum_2020}, we consider the minimum-curvature optimization problem with respect to all control point coordinates $\mathbf{z}$.
\begin{align}\label{eq:min_curvature}
    \min_{\mathbf{z}} \quad \sum_{j=1}^Mk_j^2(t)
\end{align}
where $k_1\dots k_M$ are the curvatures of the discretization points within the span of the corresponding basis function of $\mathbf{z}$. The difference from the previous work in the formulation is the optimization variable, which is the lateral movements of individual discretization points in the previous work, but is replaced with the control point placements in our work. This reduces the dimension of the decision variable from $M$, the number of discretization points, to $2S$, twice the number of control points. For Monza Circuit with 3-meter interval discretization, the dimension reduces from 1932 to 204.

We then substitute \eqref{eq:curvature} into \eqref{eq:min_curvature} and omit the constant terms in the problem. Discarding the $(t, \mathbf{z})$ notation in $T(t, \mathbf{z})$ for simplicity, we arrive at a similar QP formulation to that in Heilmeier et al. \cite{heilmeier_minimum_2020}:
\begin{align}
    \label{eq:min_quad}
    \min_{z} \quad T_x''^TP_{xx}T_x'' + T_y''^TP_{xy}T_x'' + T_y''^TP_{yy}T_y''
\end{align}
where
\begin{align*}
    P_{xx} &= \begin{bmatrix}
    \frac{(T_{y_1})'^2v_1}{((T_{x_1})'^2+(T_{y_1})'^2)^3} & 0 & \cdots & 0 \\
    0 & \ddots & \cdots & 0 \\
    \vdots & \vdots & \ddots & \vdots \\
    0 & 0 & \cdots & \frac{(T_{y_M})'^2v_M}{((T_{x_M})'^2+(T_{y_M})'^2)^3}
    \end{bmatrix}
\end{align*}

\begin{align*}
    P_{xy} &= \begin{bmatrix}
    \frac{-2(T_{x_1})'(T_{y_1})'v_1}{((T_{x_1})'^2+(T_{y_1})'^2)^3} & 0 & \cdots & 0 \\
    0 & \ddots & \cdots & 0 \\
    \vdots & \vdots & \ddots & \vdots \\
    0 & 0 & \cdots & \frac{-2(T_{x_M})'(T_{y_M})'v_M}{((T_{x_M})'^2+(T_{y_M})'^2)^3}
    \end{bmatrix}\\
    P_{yy} &= \begin{bmatrix}
    \frac{(T_{x_1})'^2v_1}{((T_{x_1})'^2+(T_{y_1})'^2)^3} & 0 & \cdots & 0 \\
    0 & \ddots & \cdots & 0 \\
    \vdots & \vdots & \ddots & \vdots \\
    0 & 0 & \cdots & \frac{(T_{x_M})'^2v_M}{((T_{x_M})'^2+(T_{y_M})'^2)^3}
    \end{bmatrix}
\end{align*}

We now need to correlate $T_x''$ and $T_y''$ with the location of the $j$-th control point $z_j = \begin{bmatrix}
    \alpha_j & \beta_j
\end{bmatrix}^T$ through the second-order derivative of the underlying B-spline basis. These lower-order bases also have a recursive closed-form solution and can be treated as constants \cite{d_boor_practical_1978}.
\begin{subequations}
    \begin{align}
        T_x''&= \sum_{i=1}^{j-1}\frac{d^2B_{i,n}(t)}{dt^2}x_i + \frac{d^2B_{j,n}(t)}{dt^2}x_j + \sum_{i=j+1}^{S}\frac{d^2B_{i,n}(t)}{dt^2}x_i \nonumber\\
        &:=F_x+\frac{d^2B_{j,n}(t)}{dt^2}x_j \label{eq:d2Tx}\\
        T_y''&= \sum_{i=1}^{j-1}\frac{d^2B_{i,n}(t)}{dt^2}y_i + \frac{d^2B_{j,n}(t)}{dt^2}y_j + \sum_{i=j+1}^{S}\frac{d^2B_{i,n}(t)}{dt^2}y_i \nonumber\\
        &:=F_y+\frac{d^2B_{j,n}(t)}{dt^2}y_j \label{eq:d2Ty}
    \end{align}
\end{subequations}
This shows that the relation between $T_x''$ and $x_j$, or $T_y''$ and $y_j$, is affine. If optimization is done with respect to all the control points at once, then $F_x$ and $F_y$ reduces to zero, since \eqref{eq:d2Tx} and \eqref{eq:d2Ty} now become
\begin{subequations}
    \begin{align}
        T_x''&= \sum_{i=1}^{S}\frac{d^2B_{i,n}(t)}{dt^2}x_i \label{eq:d2Tx_all} \\
        T_y''&= \sum_{i=1}^{S}\frac{d^2B_{i,n}(t)}{dt^2}y_i \label{eq:d2Ty_all} 
    \end{align}
\end{subequations}
In which $\{x_0,\dots,x_{S}\}$ and $\{y_0,\dots,y_{S}\}$ are all decision variables.



\begin{figure*}[h]
\centering
\includegraphics[scale=0.6,trim={0 0.5cm 0 0}]{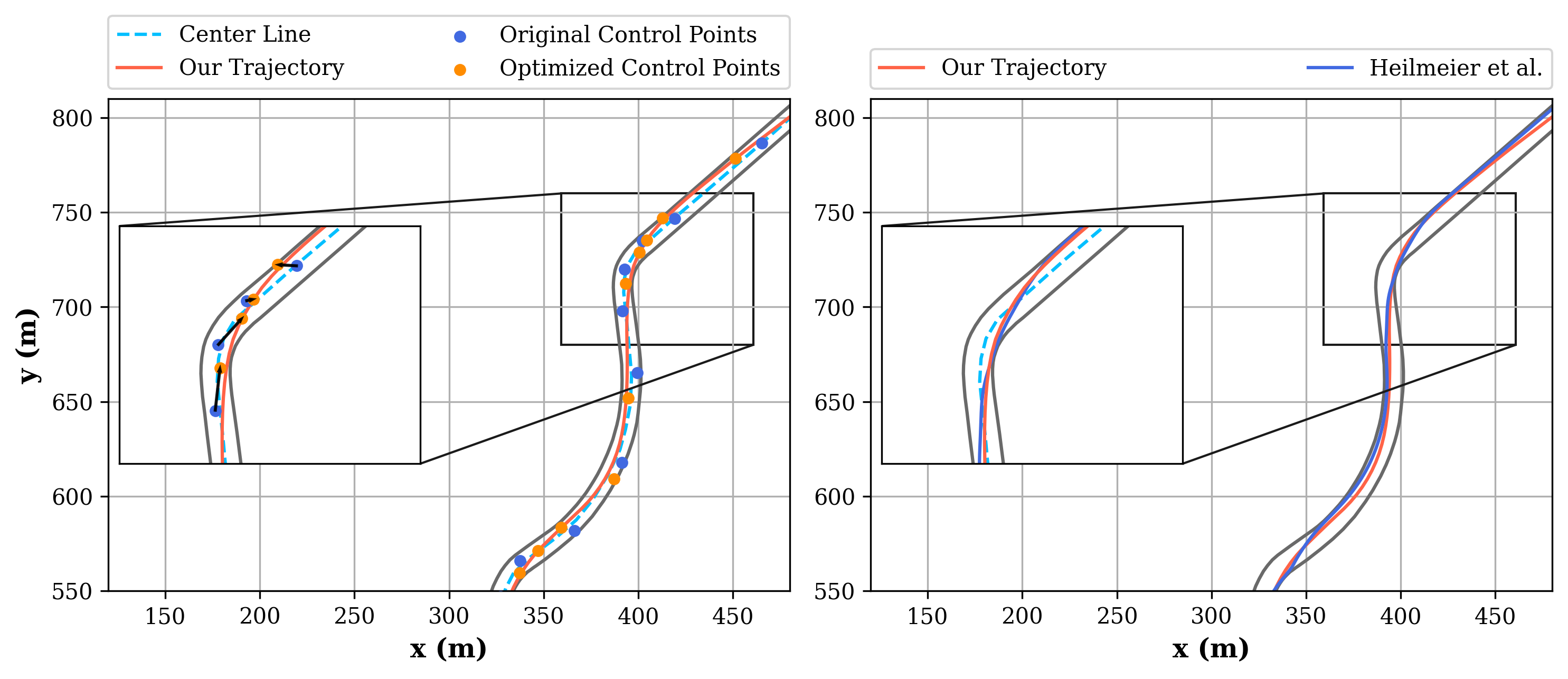}
    \caption{Optimization visualization at Monza Turn 8-10. The figure on the left visualizes the movements of the control points which shift the center line spline into a minimum-curvature spline. The figure on the right compares the optimization results of ours and \cite{heilmeier_minimum_2020}.}
 \label{fig5}
\end{figure*}


Substituting \eqref{eq:d2Tx_all} and \eqref{eq:d2Ty_all} into \eqref{eq:min_quad}, we can formulate this problem as a standard QP.
\begin{subequations}\label{eq:single_point_qp}
    \begin{align}
    & \quad \min_{\mathbf{z}} \quad \frac{1}{2} \mathbf{z}^THz+g^T\mathbf{z} \\
    \intertext{where}
    \vspace{-1cm}
    H &= B_x^T P_{xx} B_x + B_y^T P_{xy} B_x + B_y^T P_{yy} B_y \\
    g &= (F_x^T P_{xx} B_x + F_y^T P_{xy} B_y + F_y^T P_{yy} B_y) + \nonumber\\
        &\quad (B_x^T P_{xx} F_x + B_y^T P_{xy} F_x + B_y^T P_{yy} F_y)^T \label{eq:g}\\
    \intertext{and where}
    B_x &= \begin{bmatrix}
        \mathbf{B}_2 & \mathbf{0}^{M \times S}
    \end{bmatrix}^T \\
    B_y &= \begin{bmatrix}
        \mathbf{0}^{M \times S} & \mathbf{B}_2
    \end{bmatrix}^T \\
    \mathbf{B}_2 &= \begin{bmatrix}
        \frac{d^2B_{1,n}}{dt^2}(t_1) & \cdots & \frac{d^2B_{S,n}}{dt^2}(t_1) \\
        \vdots & \ddots & \vdots \\
        \frac{d^2B_{1,n}}{dt^2}(t_M) & \cdots & \frac{d^2B_{S,n}}{dt^2}(t_M)
    \end{bmatrix}
\end{align}
\end{subequations}
Note that \eqref{eq:g} is evaluated to zero when considering all the control points since $F_x=F_y=\mathbf{0}$.

\begin{figure*}[thpb]
      \centering
      \includegraphics[scale=0.6,trim={0 0.5cm 0 0}]{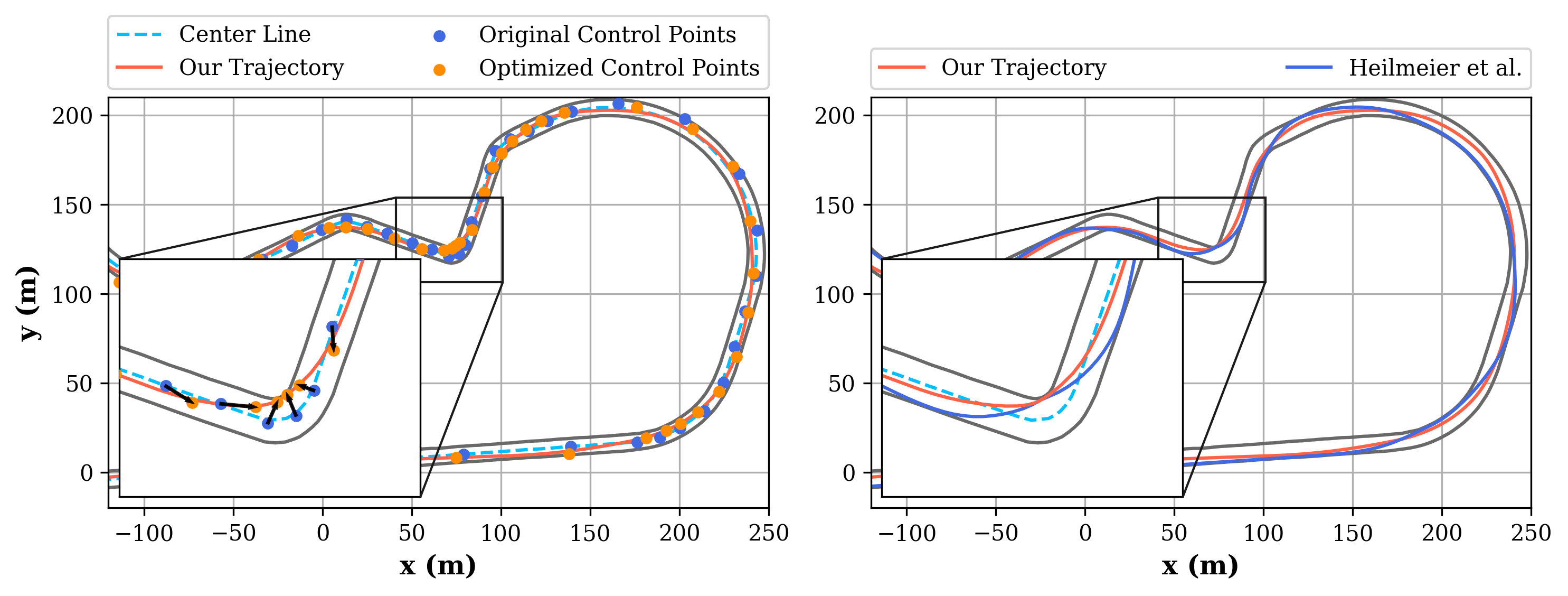}
      \caption{Optimization visualization at Putnam Road Course, Indiana, USA. The left figure shows the movement of the control points and the optimal trajectory shape as a result. The right figure compares ours and Heilmeier et al.'s trajectory through the chicane.}
      \label{fig:putnam}
\end{figure*}

Finally, we obtain the constraints for the optimization, which is the distance to the left and right boundaries. For the i-th discretization point, we use $\mathbf{l}=\begin{bmatrix}l_0 & \cdots & l_M\end{bmatrix}^T$ and $\mathbf{r}=\begin{bmatrix}r_0 & \cdots & r_M\end{bmatrix}^T$ to denote the center line's distances to the left and right boundaries. Then we use a local frenet frame approximation to find the distance between the optimized discretization point and the original center line. This is done by applying a rigid body transformation to the boundary point from the fixed map frame into the center line's local frame.

\begin{align*}
    A^{2M\times2S} &= \begin{bmatrix}
    \mathbf{B}
    & \mathbf{0^{M\times S}} \\
    \mathbf{0^{M\times S}} &
    \mathbf{B}
    \end{bmatrix}
\end{align*}

\begin{align*}
    R^{M\times2M} &= \begin{bmatrix}
    \begin{matrix}
        \cos \theta_1 & 0 & 0 \\
        0 & \ddots & 0 \\
        0 & 0 & \cos \theta_M
    \end{matrix} &
    \begin{matrix}
        \sin \theta_1 & 0 & 0 \\
        0 & \ddots & 0 \\
        0 & 0 & \sin \theta_M
    \end{matrix} &
    \end{bmatrix} \\
    \intertext{where}
    \mathbf{B} &= \begin{bmatrix}
        B_{1,n}(t_1) & \cdots & B_{S,n}(t_1) \\
        \vdots & \ddots & \vdots \\
        B_{1,n}(t_M) & \cdots & B_{S,n}(t_M)
    \end{bmatrix}
\end{align*}
and where $\theta_i \quad \forall i \in \{1, \dots, M\}$ denotes the orientation of the center line at discretization point $p_i$. The linear constraint is therefore
\begin{align*}
    \mathbf{r} \le RA\mathbf{z} \le \mathbf{l}
\end{align*}
Breaking down this equation, $A\mathbf{z}$ computes the discretization coordinates from the control points, which is rotated by $R$ to compute its projection on the left or right side of the center line.

Now we have all the components for our constrained QP problem, which we can solve with qpOASES \cite{ferreau_qpoases_2014}. 
\begin{align}
    \min_{\mathbf{z}} \quad &\frac{1}{2} \mathbf{z}^TH\mathbf{z}+g^T\mathbf{z} \nonumber\\
    \text{s.t.} \quad &\mathbf{r} \le RA\mathbf{z} \le \mathbf{l}
\end{align}

\section{RESULTS} \label{results}

Before presenting the results, it is beneficial to discuss the evaluation metrics for OTO methods. Since this is an open-loop offline method, it is insufficient to show the end trajectory achieved by a vehicle driving on the track, since there are other processes down the pipeline such as online dynamic planning and optimal control calculations. These modules also have a significant impact on the outcome of the experiment. Therefore, we will directly compare the OTO result with the current state-of-the-art method in minimum-curvature optimization in \cite{heilmeier_minimum_2020} to see if our method has achieved our objectives, which is to offer an alternative minimum-curvature optimization formulation for autonomous racing that uses significantly fewer decision variables, guarantees continuity and still offers a trajectory comparable to previous work.

To this end, we will perform minimum-curvature optimization using our formulation and Heilmeier et al. formulation \cite{heilmeier_minimum_2020}. The discretization intervals are the same in both experiments. We then simulate lap times with the QSS simulation method described in III.A to show that we have reached comparable results with the previous work.

We first compare the overall lap time performance of our method with the reference center line and Heilmeier et al. \cite{heilmeier_minimum_2020} in Table \ref{table-compare}. The optimization algorithms produce 7.65\% (ours) and 8.94\% (Heilmeier et al.) lap time reduction, respectively, offering comparable maximum velocity and acceleration profiles. The percentage difference of the metrics is shown in the right column, which suggests that we achieved a level of optimality similar to \cite{heilmeier_minimum_2020}. However, our result produces a $C^2$ continuous B-spline curve that can be easily re-discretized and even artificially manipulated afterwards by dragging the control points, which is a very useful feature in the developmental stage of autonomous racing, whereas \cite{heilmeier_minimum_2020} only produces a discretized trajectory with no guarantee of continuity level due to its discrete decision variables. We also see a significant reduction in the dimension of the decision variable. Our QP problem has 204 decision variables (102 control points), down from 1932 using \cite{heilmeier_minimum_2020} with a 3 \si{\meter} discretization interval. The resulting QP computation time is 3.8 \si{\milli\second}, down from 8.225 \si{\second} using \cite{heilmeier_minimum_2020}. This shows that our method not only gives a more compact formulation that captures the essence of the minimum-curvature problem, but also enables the possibility of adopting such method in online planning on a very long track.

Looking turn by turn, Fig. \ref{fig5} zooms in on the optimization at turns 8-10. The control points of the center-line spline are shifted to form a minimum curvature trajectory. For turn 7 in the zoomed-in section, we are able to optimize the full trajectory with only four control points.


\begin{table}[h]
\centering
\smallskip
\smallskip
\caption{Simulation Results on Monza} \label{table-compare}
\begin{tabular}{l|rrrr}\toprule
           & \textbf{Center} & \textbf{Heilmeier} & \textbf{Ours}  & \textbf{Delta} \\
           & \textbf{Line} & \textbf{et al.} & ~ & \textbf{(\%)} \\ \midrule
\textbf{Lap Time $(s)$}              & 131.33               & 119.59                    & 121.28  &-1.4      \\ 
\textbf{Ave Speed $(m/s)$}           & 43.72                & 48.60                     & 47.73   &1.8      \\ 
\textbf{Max Speed $(m/s)$}           & 93.69                & 95.12                     & 94.33   &0.8      \\ 
\textbf{Min Speed $(m/s)$}           & 6.04                 & 10.51                     & 9.44    &11.3      \\
\textbf{Max Lat G $(m/s^{-2})$}      & 13.96                & 13.78                     & 14.48   &-4.8      \\ 
\textbf{Max Throttling $(m/s^{-2})$} & 9.74                 & 9.62                      & 9.68    &-0.6      \\ 
\textbf{Max Braking $(m/s^{-2})$}    & -19.87               & -19.27                    & -19.85  &-2.9      \\ \bottomrule
\end{tabular}
\end{table}

We then move on to a different race track and apply the same OTO method  to see how our method transfers to a new track. Fig. \ref{fig:putnam} visualizes the optimization results done on Putnam Road Course, Indiana, USA, which is a test track of the Indy Autonomous Challenge. The zoomed-in portion of the graph visualizes the movement of control points which leads to a curvature-optimal trajectory. Table \ref{tab:putnam_metrics} shows metrics similar to \ref{table-compare}, in which comparable lap time reduction of 13.9\% and 15.67\% are respectively achieved by \cite{heilmeier_minimum_2020} and our work. The difference in the simulation metrics between ours and \cite{heilmeier_minimum_2020} remains close, suggesting that the outcome of the optimization is very comparable.
 
\begin{table}[]
\centering
\caption{Simulation Results on Putnam Road Course}
\label{tab:putnam_metrics}
\begin{tabular}{l|rrrr}\toprule
           & \textbf{Center} & \textbf{Heilmeier} & \textbf{Ours}  & \textbf{Delta} \\
           & \textbf{Line} & \textbf{et al.} & ~ & \textbf{(\%)} \\ \midrule
\textbf{Lap Time $(s)$}              & 79.63                & 68.55                     & 66.91  & 2.4       \\
\textbf{Ave Speed $(m/s)$}           & 36.36                & 41.57                     & 42.75  &-2.7       \\
\textbf{Max Speed $(m/s)$}           & 73.93                & 86.54                     & 88.07  &-1.7       \\
\textbf{Min Speed $(m/s)$}           & 8.02                 & 13.83                     & 12.0   &15.2        \\
\textbf{Max Lat G $(m/s^{-2})$}      & 15.0                 & 15.0                      & 15.0   &0.0        \\
\textbf{Max Throttling $(m/s^{-2})$} & 9.75                 & 9.18                      & 9.52   &-3.6        \\
\textbf{Max Braking $(m/s^{-2})$}    & -16.95               & -17.84                    & -18.22 &-2.1      \\ \bottomrule
\end{tabular}
\end{table}

\section{CONCLUSIONS}

We present a B-spline OTO method for autonomous racing which solves a minimum-curvature optimization problem. Compared to previous works which only output a discretized trajectory, this work outputs a fully parameterized trajectory of $C^2$ continuity, which ensures a smooth control profile for high-speed vehicle handling. The algorithm also considers the data scarcity of early-stage autonomous motorsports development and requires minimum vehicle dynamics data. Compared to previous work \cite{heilmeier_minimum_2020}, the dimension of the problem is significantly reduced from thousands of discretization points down to a few dozens of control knot points. The problem computation time is also drastically reduced as a result. This enables future work to explore the online application of minimum-curvature OTO for autonomous racing.


\addtolength{\textheight}{-12cm}   





\newpage
\bibliographystyle{plain}
\bibliography{trajectory_optimization}

\end{document}